\documentclass[review]{elsarticle}

\usepackage{lineno,hyperref}
\usepackage{amsmath,amssymb,amsfonts}
\usepackage{algorithmic}
\usepackage{algorithm}
\usepackage{array}
\usepackage{textcomp}
\usepackage{stfloats}
\usepackage{url}
\usepackage{verbatim}
\usepackage{graphicx}
\usepackage{multirow}
\usepackage{booktabs}

\newcommand{\ie}{\textit{i}.\textit{e}.}

\journal{Journal of \LaTeX\ Templates}

\begin{document}

\begin{frontmatter}

\title{Image to Pseudo-Episode: \\Boosting Few-Shot Segmentation by Unlabeled Data}





\author[inst1,inst2,inst4]{Jie Zhang}
\cortext[mycorrespondingauthor]{Corresponding author}
\ead{zhangjie@ict.ac.cn}

\author[inst1,inst2]{Yuhan Li}
\ead{yuhan.li@vipl.ict.ac.cn}

\author[inst1,inst2]{Yude Wang}
\ead{yude.wang@vipl.ict.ac.cn}

\author[inst3]{Stephen Lin}
\ead{stevelin@microsoft.com}

\author[inst1,inst2]{Shiguang Shan\corref{mycorrespondingauthor}}
\ead{sgshan@ict.ac.cn}

\address[inst1]{Key Lab of Intelligent Information Processing of Chinese Academy of Sciences (CAS), Institute of Computing Technology, CAS, Beijing, 100190, China}
\address[inst2]{University of Chinese Academy of Sciences, Beijing, 100049, China}
\address[inst3]{Microsoft Research Asia, Beijing, 100080, China}
\address[inst4]{Institute of Intelligent Computing Technology, CAS, Suzhou, 215124, China}

\begin{abstract}
Few-shot segmentation (FSS) aims to train a model which can segment the object from novel classes with a few labeled samples. The insufficient generalization ability of models leads to unsatisfactory performance when the models lack enough labeled data from the novel classes. Considering that there are abundant unlabeled data available, it is promising to improve the generalization ability by exploiting these various data. For leveraging unlabeled data, we propose a novel method, named Image to Pseudo-Episode (IPE), to generate pseudo-episodes from unlabeled data. Specifically, our method contains two modules, i.e., the pseudo-label generation module and the episode generation module. The former module generates pseudo-labels from unlabeled images by the spectral clustering algorithm, and the latter module generates pseudo-episodes from pseudo-labeled images by data augmentation methods. Extensive experiments on PASCAL-$5^i$ and COCO-$20^i$ demonstrate that our method achieves the state-of-the-art performance for FSS.
\end{abstract}


\end{frontmatter}

\section{Introduction}
Deep learning has made tremendous progress in recent years with various neural networks performing well on many computer vision tasks, such as classification~\cite{he2016deep}, object detection~\cite{he2017mask}, and semantic segmentation~\cite{chen2018encoder}. Although neural networks are effective under normal circumstances, their demands on large amounts of annotated data limit their application, as data labeling is labor-intensive, especially for dense prediction tasks like semantic segmentation. To ease the dependence on large annotated data, there exist several approaches: 1) unsupervised pre-training~\cite{chen2020simple,wang2021dense} aims to provide a good initialization for downstream tasks while getting rid of the reliance on annotation during pre-training with unlabeled data; 2) weakly supervised~\cite{kolesnikov2016seed,saleh2016built} segmentation attempts to tackle the segmentation problem with weakly annotated data; 3) few-shot segmentation (FSS) emphasizes on how to effectively transfer the features learned from base classes to the novel classes for obtaining a good segmentation model with a few labeled data of novel classes. In this paper, we mainly focus on few-shot segmentation.

\begin{figure*}[t]
 \centering
 \includegraphics[width=1\linewidth]{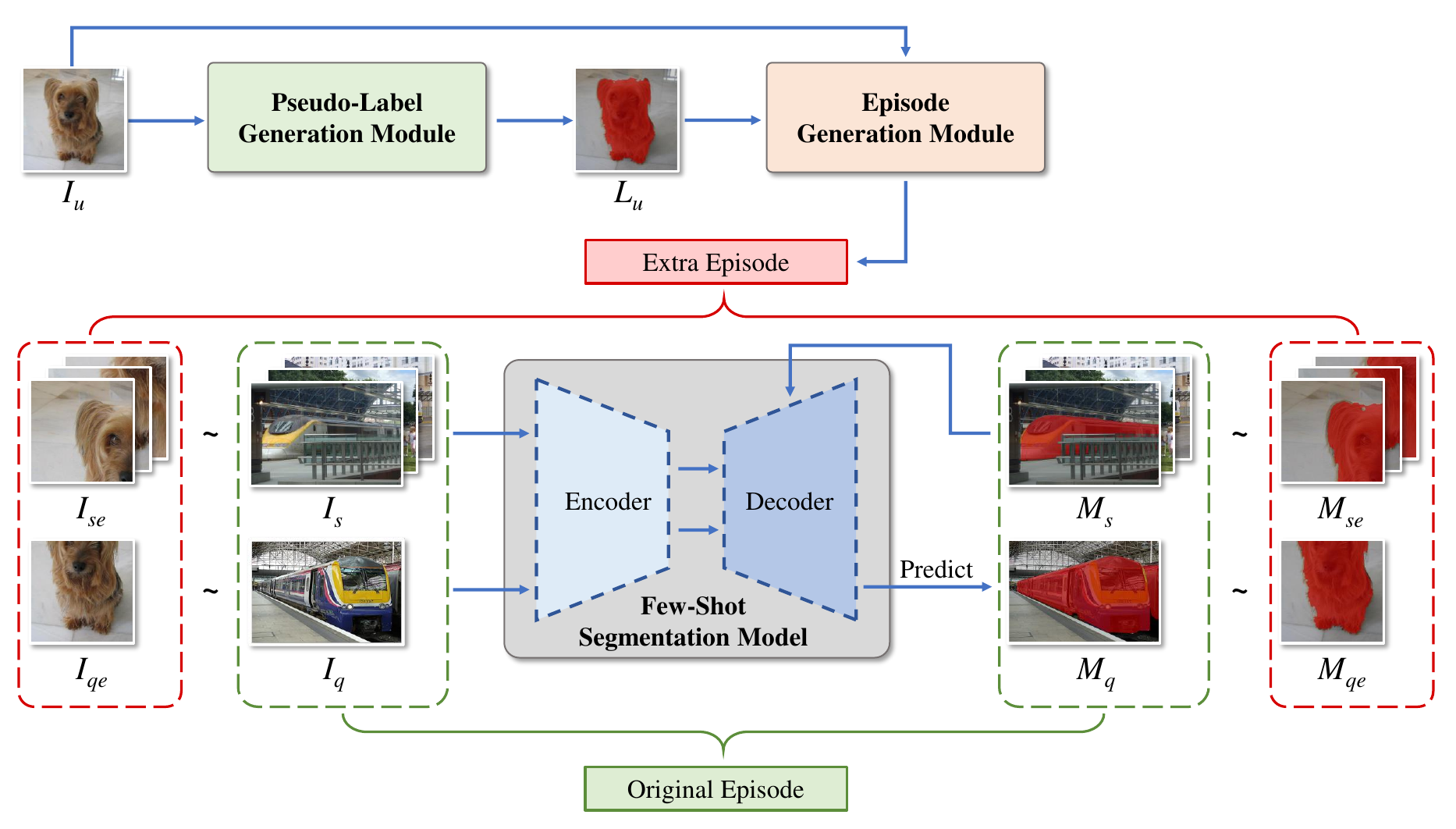}
 \caption{Overview of our `Image to Pseudo-Episode' method and its combination with other few-shot segmentation methods. $I_u$ represents the unlabeled image; $L_u$ represents the pseudo-label; $I_s$, $M_s$, $I_q$ and $M_q$ represent support images, support masks, query image and query mask respectively; the subscript $e$ represents `extra', means its owner is from an unlabeled image.}
 \label{fig:fss_model}
\end{figure*}

Mainstream FSS methods~\cite{zhang2019canet,min2021hypercorrelation,tian2020prior,yang2020prototype,li2021adaptive, shi2022dense, johnander2022dense, zhang2022feature} have an encoder-decoder structure as illustrated in Figure~\ref{fig:fss_model}. The weight-shared encoder, which is usually a convolutional neural network (e.g., VGG~\cite{simonyan2014very} and ResNet~\cite{he2016deep}), extracts the features from the query image as well as the support images. Then these features are fed to the decoder together with the support masks. Finally, the decoder mines the correlation between the query and support data to predict the segmentation mask of the query image. Since a few data of novel classes are available in the FSS problem, a good initialization of the encoder is crucial. This is the reason why almost all FSS methods~\cite{min2021hypercorrelation,shi2022dense,tian2020prior,li2021adaptive} require an ImageNet~\cite{deng2009imagenet} pre-trained encoder; otherwise, they cannot achieve good performance. Besides, recent research has found that limited base data impairs the feature representation learning and usually causes serious over-fitting problems, so many methods turn to freezing the encoder, which can preserve the diversity of features as much as possible. But, unlike the encoder, the decoder only receives information from base data, which seriously restrains the generalization ability of the whole model. Currently, existing works mainly focus on modifying the structure of the decoder, which does not fundamentally solve the over-fitting problem. We believe that introducing more information from extra data is the key to addressing this issue. Since the cost of collecting a large number of finely labeled images is high, leveraging unlabeled data to boost the generalization ability of the model is an economical and promising solution.

Under the FSS setting, the classes of data are divided into base classes and novel classes without overlap. The data from base classes are utilized for training the model and the data from novel classes are employed for testing. The process of training and test are both episodical~\cite{shaban2017one}, which means a few annotated support images and a query image are fed into the model in pairs to generate the prediction for the query image. As shown in Figure~\ref{fig:fss_model}, the set of query data and support data is called an `episode', and all data in an episode belong to the same class. The episodical setting implies that FSS models need to find similar regions between support and query images. Since the affinity map which consists of the cosine-similarity between each pixel pair of the feature map naturally contains abundant pixel-wise similarity information, exploring the affinity map of extra unlabeled images could be a feasible approach for gaining extra supervision. Inspired by this idea, we propose a novel method, named Image to Pseudo-Episode (IPE), to leverage large unlabeled data to improve the performance of FSS methods. As shown in Figure~\ref{fig:fss_model}, our method consists of two modules, the Pseudo-label Generation Module (PGM) and the Episode Generation Module (EGM). The PGM utilizes the spectral clustering algorithm~\cite{fiedler1973algebraic} and dense conditional random field algorithm~\cite{krahenbuhl2011efficient} to generate pseudo-labels from the affinity map of unlabeled images. And the EGM utilizes data augmentation methods to generate episodes from pseudo-labels and images, which is inspired by contrastive learning. Since the usage of these extra episodes is consistent with the original episodes, the IPE is suitable for most FSS models without modifying their structure. Through the above process, IPE successfully brings rich information from abundant unlabeled data to improve FSS models.

Recent studies~\cite{he2020momentum,grill2020bootstrap,wang2021dense} have demonstrated that unsupervisedly pre-trained models can generate relatively complete features and have achieved better performance than supervisedly pre-trained models on many downstream tasks (e.g., image recognition, object detection and semantic segmentation). But the performance of FSS models degenerates if we simply replace the encoder initialization from the model supervisedly pre-trained on ImageNet ~\cite{deng2009imagenet} to the unsupervised one, as shown in Table \ref{tab:pascal1} and \ref{tab:coco1}. This phenomenon suggests that the supervisedly pre-trained model on ImageNet plays an important role in existing FSS methods, which is probably because of the overlap between novel classes and classes in ImageNet. In view of the above, we believe that tackling the FSS problem with the unsupervisedly pre-trained model is more reasonable, leading us to push the frontier of the area by leveraging unlabeled data to boost Few-Shot segmentation.

Extensive experiments on PASCAL-$5^i$~\cite{shaban2017one} and COCO-$20^i$~\cite{lin2014microsoft,nguyen2019feature} demonstrate that our method boosts the performance of mainstream FSS models (i.e., PFENet~\cite{tian2020prior}, BAM~\cite{lang2022learning}) under both the supervisedly and unsupervisedly pre-trained settings. The performance of PFENet combined with our method under the unsupervisedly pre-trained setting even exceeds the original PFENet with supervised initialization. All these experiments demonstrate the effectiveness of our method from different aspects.

\section{Related work}

\subsection{Semantic segmentation}
Semantic segmentation is a fundamental computer vision task, whose goal is to recognize the category of each pixel of the input image. Prevalent segmentation methods~\cite{lin2017refinenet,zhao2017pyramid,ronneberger2015u,chen2017rethinking,orvsic2021efficient} typically contain an encoder-decoder structure, where the encoder extracts the feature maps and the decoder builds the pixel-wise prediction by these maps. This architecture is derived from FCN~\cite{long2015fully}, which proposes the 1x1 Conv for pixel-wise classification. After FCN, Chen et al.~\cite{chen2017deeplab} propose the dilated convolution and Zhao et al.~\cite{zhao2017pyramid} propose the pyramid pooling. Both of them enlarge the receptive field of the model and achieve significant improvements for segmentation. Although current semantic segmentation methods are effective, they have the same defect: a large amount of well-annotated data is required to build a reliable model and it is hard to adapt to tackle novel classes with only a few samples.

\subsection{Few-shot Segmentation}
Few-shot learning aims at using abundant base samples to train a model that can achieve good performance on novel classes with very limited data. Early attempts focus on few-shot image classification, and many classic methods~\cite{chen2019closer,snell2017prototypical,garcia2017few,koch2015siamese,oreshkin2018tadam,vinyals2016matching,huang2021local} are proposed to tackle this challenging task. As research progressed, researchers began to consider addressing semantic segmentation tasks under the few-shot setting and were deeply inspired by the few-shot classification methods. 

OSLSM~\cite{shaban2017one}, the first work on the Few-Shot Segmentation (FSS) problem,  follows the episode setting of ~\cite{vinyals2016matching} and slightly modifies it to fit the FSS problem. Also in OSLSM, the two-branch network which consists of a support branch and a query branch is proposed to extract features from several support images and one query image respectively. Then these features are merged to generate the prediction of the query image. After OSLSM, FSS has become a popular emerging research direction, then many excellent methods have been proposed one after another.

Inspired by~\cite{snell2017prototypical}, PL~\cite{dong2018few} proposes the prototypical network and utilizes global average pooling (GAP)~\cite{lin2013network} to generate prototypes from support images, where these prototypes support the model to predict the mask of the query image. SG-One~\cite{zhang2020sg} argues that generating a prototype by masked average pooling is better than GAP since it relieves the influences from background noises and forces the model to focus on learning object features. Currently, the two-branch architecture, prototype architecture and masked average pooling have become the most popular designs in tackling FSS problem.

DGPNet~\cite{johnander2022dense} proposes a novel method utilizing dense gaussian process regression to tackle FSS problem, and achieves the higher accuracy. SVF~\cite{sun2022singular} introduces a novel method for fine-tuning the backbone by selectively adjusting only the singular values in its weights. This approach stands in contrast to the common practice in most FSS methods, where the backbone is frozen to mitigate over-fitting.

With the development of Vision Transformers, FSS methods based on the Transformer architecture~\cite{zhang2022feature, shi2022dense,zhang2021few} have also emerged. FPTrans~\cite{zhang2022feature} stands out as one representative work. It points out that simply replacing the CNN-based encoder with a Transformer does not lead to accuracy improvement for FSS. To address this, they propose an effective FSS method that fully utilizes the Transformer structure. Besides DCAMA~\cite{shi2022dense} proposes using cross-query-and-support attention to weight the pixel-wise additive aggregation of support mask values for predicting the query mask, which is also implemented in the form of Transformer. The effectiveness of these methods shows that the Transformer is also a promising structure to tackle the FSS problem.

Recently, Tian et al.~\cite{tian2020prior} propose PFENet that utilizes training-free prior masks to enhance its generalization and design a feature enrichment module (FEM) for overcoming spatial inconsistency of objects. Due to the excellent design of FEM, several subsequent works have inherited this design or used it as a baseline, such as ASGNet~\cite{li2021adaptive} and SCL~\cite{zhang2021self}. BAM~\cite{lang2022learning} applies an additional base learner to recognize the base classes from input images to guide the prediction of novel classes, and achieves the sota results at that time. we adopt PFENet
and BAM as our baselines.

\subsection{Self-supervised pre-training}
Self-supervised learning~\cite{he2020momentum,grill2020bootstrap,chen2020simple,wang2021dense}, which aims at getting rid of the dependence on a large amount of labeled data, has achieved great success in several tasks (e.g., image recognition, object detection and semantic segmentation). Contrastive learning, an important and popular branch of self-supervised learning, learns similar representations of positive pairs and dissimilar representations of negative pairs, where the positive pair is formed from the same image through different augmentation pipelines and the negative pairs are formed from different images.

To better adapt to downstream dense prediction tasks, several works~\cite{wang2021dense,xie2021detco} modify the generation process of positive/negative pairs or the contrastive loss. DenseCL~\cite{wang2021dense} designs an effective contrastive learning method that directly works at the pixel level and achieves amazing performance on the downstream semantic segmentation task.

\section{Proposed Approach}
\subsection{Overview}
FSS aims at training a segmentation model on abundant annotated data, which can segment the object from novel classes ${C}_N$ with only a few labeled samples.
The "novel" means that these classes are not included in the base classes ${C}_B$ that are usually used for training. In other words, ${C}_B \cap {C}_N = \varnothing $. 
Formally, we need to design a model $\mathcal{F}$ that generates the pixel-wise prediction $P$ of each class from input data $X$, \ie, $P = \mathcal{F}(X)$. 
Our goal is to optimize the parameters $\boldsymbol{\theta}_\mathcal{F}$ of $\mathcal{F}$ that can properly distinguish the classes for each pixel during the test. 
In segmentation tasks, the cross-entropy loss is usually employed as the loss function, which is defined as
\begin{equation}
\mathcal{L}=-\frac{1}{N\cdot\left|\mathrm{\Psi}\right|} \sum_{i=1}^{N} \sum_{j \in \mathrm{\Psi}} {y}_{ij} \log \left(p_{ij}\right), \label{function:CELoss}
\end{equation}
where $\mathrm{\Psi}$ is the set of all pixels, $N$ is the number of all classes, $p_{ij}$ is the probability of pixel $j$ belonging to class $i$ predicted by model $\mathcal{F}$ and $y_{ij}$ denotes the groundtruth of the pixel-level segmentation mask. If the pixel $j$ actually belongs to the class $i$, $y_{ij}$ will be $1$; otherwise it will be $0$.

\begin{figure*}[t]
 \centering
 \includegraphics[width=1.0\linewidth]{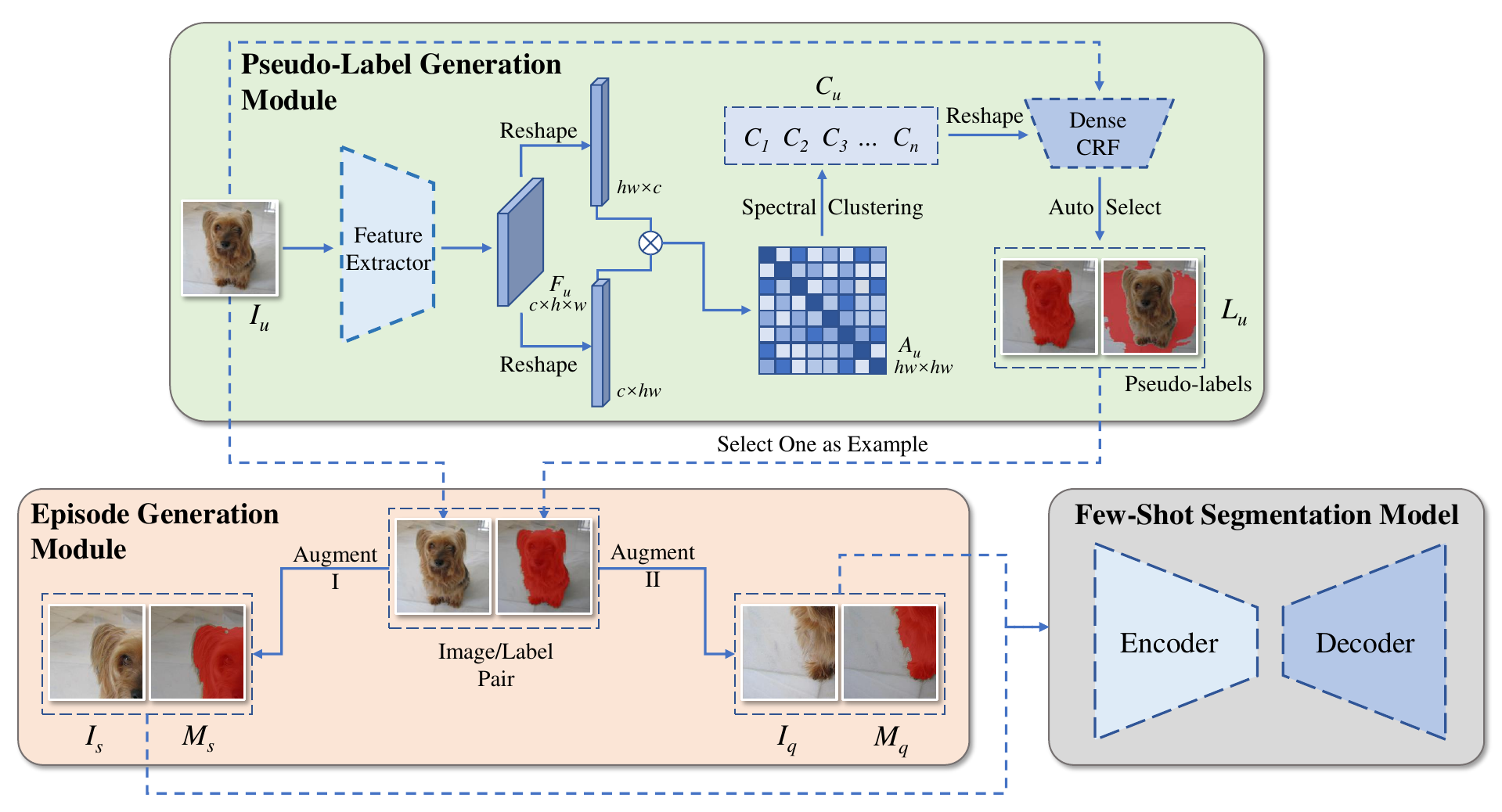}
 \caption{Visual illustration of IPE under 1-shot scenario. $I_u$ represents the unlabeled image; $L_u$ represents the pseudo-label; $I_{se}$, $M_{se}$, $I_{qe}$ and $M_{qe}$ represent extra support image, extra support mask, extra query image and extra query mask respectively; $c$, $h$ and $w$ are dimensions which represent channel, height and width respectively; $C_1$ to $C_n$ represents the clusters in clustering result.}
 \label{fig:IPE}
\end{figure*}

Most FSS methods follow the 1-way setting and the episode paradigm which are proposed in OSLSM~\cite{shaban2017one}. 
The 1-way setting means there is only one target class to be predicted, which constrains $N$ in Equation ~\ref{function:CELoss} to be equal to $2$ (the other class is the background). 
The episode paradigm is shown in Figure~\ref{fig:fss_model}, where each episode consists of two sets, \ie, a query set $Q$ and a support set $S$. 
The query set $Q$ is a pair of the query image $I_q$ and the corresponding groundtruth mask $M_q$, while the support set $S$ consists of $K$ support images $\{I_s^1, I_s^2,\dots, I_s^K\}$ and the corresponding groundtruth masks $\{M_s^1, M_s^2,\dots, M_s^K\}$ under the $K$-shot scenario.
Then, the FSS model $\mathcal{F}$ aims to predict the segmentation mask of the query image $I_q$ by taking $\{ I_s^1, I_s^2,\dots, I_s^K, M_s^1, M_s^2,\dots, M_s^K , I_q \}$ as input.

Most FSS methods have an encoder-decoder structure. The encoder extracts features from $I_q$ and $\{ I_s^1, I_s^2,\dots, I_s^K\}$, and the decoder utilizes these features together with $\{ M_s^1, M_s^2,\dots, M_s^K\}$ to infer the segmentation mask of $I_q$: 
\begin{equation}
\mathcal{F}(X) \triangleq \mathcal{D}(\mathcal{E}(I_q, I_s^1, I_s^2,\dots, I_s^K ), M_s^1, M_s^2,\dots, M_s^K ),
\end{equation}
where $\mathcal{D}$ refers to the decoder network, $\mathcal{E}$ refers to the encoder network which is usually initialized by a supervisedly pre-trained model on ImageNet~\cite{deng2009imagenet}. Since most recent FSS methods freeze the encoder, each model mainly differs in the decoder. We focus on improving the decoder $\mathcal{D}$ by leveraging the large-scale unlabeled data. Moreover, we explore the potential of employing an unsupervisedly pre-trained model to initialize the encoder $\mathcal{E}$, which we believe is a more reasonable setting for few shot segmentation tasks. Detailed results and discussions can be found in Section~\ref{section:result}.

\subsection{Unlabeled Image to Pseudo-Episode}
For leveraging extra unlabeled data to boost the generalization ability of the decoder, one roadmap is to modify the structure of the decoder to fit with extra unlabeled data, which is difficult to achieve and also lacks scalability to other models. The other is to exploit pseudo-labels of the unlabeled data to extend episodical inputs. We resort to the latter roadmap by proposing Image to Pseudo-Episode (IPE) module $\mathcal{P}$. Specifically, the module $\mathcal{P}$ consists of two modules, \ie, the Pseudo-label Generation Module (PGM) $\mathcal{P}_1$ and Episode Generation Module (EGM) $\mathcal{P}_2$, where the first module is for generating pseudo-labels from unlabeled data, and the second module is for generating pseudo-episodes from pseudo-labeled data:
\begin{equation}
\{{I_{qe}},{M_{qe}},{I_{se}^1},{I_{se}^2},\dots,{I_{se}^K},{M_{se}^1},{M_{se}^2},\dots,{M_{se}^k}\} = \mathcal{P}_2(I_u, \mathcal{P}_1(I_u)),
\label{function:IPE}
\end{equation}
where $I_u$ is one unlabeled image; the left part of Equation~\ref{function:IPE}  denotes the input data batch for $K$-shot episode training. ${I_{se}}$ and ${M_{se}}$ refer to the support image and the corresponding mask, respectively. ${I_{qe}}$ and ${M_{qe}}$ refer to the query image and the corresponding mask, respectively.

\subsection{Pseudo-label Generation Module}
\label{section:PGM}

\begin{algorithm}[t]
    \label{algorithm:PGM}
	\caption{Pseudo-label Generation} 
	\begin{algorithmic}[1]
	\REQUIRE $I_u$: unlabeled image
		\STATE extract the feature map $F_u$ of $I_u$
		\STATE calculate the affinity map $A_u$ by $F_u$
		\FOR{$k=m$ to $n$}
		    \STATE generate the spectral clustering result $C_k$ of $A_u$ with the cluster number $k$
		    \STATE calculate the Calinski-Harabasz score $S_{CH}^i$ of $C_k$
		\ENDFOR
		\STATE select the {$C_{best}$} which has the highest $S_{CH}$ from $\{C_m, \dots, C_n\}$
		\STATE generate the refined {$C_{best}$} via dense-CRF on $I_u$ and {$C_{best}$}
		\STATE select the top-$t$ pseudo-labels $\{L_1, \dots, L_t\}$ from the $k$ clusters of {$C_{best}$} with the smallest average distance to the center of $I_u$
	\ENSURE $\{L_1, \dots, L_t\}$: the pseudo-label set of $I_u$
	\end{algorithmic}  
\end{algorithm}

In this section, we introduce the Pseudo-label Generation Module (PGM) in detail, which aims to generate pseudo-labels from unlabeled data. As illustrated in Figure~\ref{fig:IPE} and Algorithm~\ref{algorithm:PGM}, the PGM firstly utilizes a feature extractor $\mathcal{H}$, which is pre-trained with unlabeled data (ImageNet-1K), to generate high-level feature $F_u$ from an unlabeled image $I_u$:
\begin{equation}
F_u = \mathcal{H}(I_u),
\end{equation}
where $F_u\in \mathbb{R}^{c\times h\times w}$, $c$ is the number of channels, $h$ and $w$ is the height and width of the feature map.

Then, we reshape the feature $F_u$ to $F_u^{'} \in \mathbb{R}^{c\times hw}$, and calculate affinity map $A_u \in \mathbb{R}^{hw\times hw}$ by the cosine-similarity between each pixel-pair in the feature map $F_u^{'}$:

\begin{equation}
A_{u}^{\left(i, j\right)} = \cos \left(f_{i}, f_{j}\right),\quad i,j\in {1, 2, \dots, hw},
\end{equation}

\begin{equation}
\cos \left(f_{i}, f_{j}\right)=\frac{f_{i}^{\top}\cdot f_{j}}{\left\|f_{i}\right\|\cdot\left\|f_{j}\right\|},\quad i,j\in {1, 2, \dots, hw},
\end{equation}
{where $f_{i}$ and $f_{j}$ are the feature vectors corresponding to pixels $i,j$, respectively.} 

To find regions with high internal similarity, we perform spectral clustering~\cite{fiedler1973algebraic} on the affinity map $A_u$. Then we choose the clustering result {$C_{best}$} with the highest Calinski-Harabasz score~\cite{calinski1974dendrite} from $\{C_{m}, \dots, C_{n}\}$, where the $C_{m}$ and $C_{n}$ are the clustering result with the minimal and maximum number of clusters, respectively. The Calinski-Harabasz score is defined as
\begin{equation}
S_{CH}=\frac{\operatorname{tr}\left(B_{k}\right)}{\operatorname{tr}\left(W_{k}\right)} \times \frac{n_{\psi}-k}{k-1},
\end{equation}
where $n_{\psi}$ is the number of all points, $k$ is the total number of clusters, $\operatorname{tr}(*)$ represents the trace of a matrix, $B_{k}$ is the between-class scatter matrix and $W_{k}$ is the within-class scatter matrix. $B_{k}$ and $W_{k}$ are defined as below:
\begin{equation}
W_{k}=\sum_{q=1}^{k} \sum_{x \in C_{q}}\left(x-c_{q}\right)\left(x-c_{q}\right)^{\top},
\end{equation}
\begin{equation}
B_{k}=\sum_{q=1}^{k} n_{q}\left(c_{q}-c_{\psi}\right)\left(c_{q}-c_{\psi}\right)^{\top},
\end{equation}
where $k$ is the total number of clusters, $c_{q}$ is the center of the $q$-th cluster, $C_{q}$ is the set of samples in the $q$-th cluster, $n_{q}$ is the number of samples in set $C_{q}$, and $c_{\psi}$ is the center of all samples.

Then, we perform the dense-CRF on $I_u$ with the {$C_{best}$} as the initial probability map to refine the boundary of each cluster. After that, we select the top-$t$ clusters with the smallest average distance to the center of the image $I_u$ as the pseudo-labels. 
The average distance is calculated as
\begin{equation}
D_q = \frac{\sum_{x \in q}{{\left\|x - c\right\|}_2}}{n_q},
\end{equation}
where $n_{q}$ is the number of samples in the $q$-th cluster, and $c$ is the center of the image $I_u$. The $t$ pseudo-labels $\{L_1, \dots, L_t\}$ generated by PGM will be sent to EGM, and all the results will be used as extra inputs for training. 
The reason why we choose multiple clusters lies in two aspects: 1) there may be multiple different foreground objects in a single image; 2) backgrounds with high internal similarity still help the FSS model learn how to find similar regions.

\subsection{Episode Generation Module}
After obtaining pseudo-labels for the unlabeled data, we further design Episode Generation Module (EGM) to generate extra episodes for training. As illustrated in Figure~\ref{fig:IPE}, we conduct image transformations to generate different views of the input image and the corresponding mask. Specifically, we conduct a set $\mathcal{T}$, which consists of several image transformations. Under the $K$-shot setting, we randomly draw $K + 1$ subset $\mathcal{T}_1,\dots,\mathcal{T}_{K + 1}$ from $\mathcal{T}$ to generate extra episodes as below:
\begin{equation}
\{{I_i}, {M_i}\} = \mathcal{T}_i(I_u, L_u),\quad  i\in\{1,\dots,K + 1\},
\end{equation}
where $L_u$ denotes pseudo-label of unlabeled image $I_u$ we generated from PGM. $\{{I_i}, {M_i}\}$ represents the $i$-th transformed image and its label. We randomly choose one pair as the extra query set $Q$ and the other $K$ pairs as the extra support set {$S$}.

\section{Experiments}

To verify the effectiveness of our method, we conduct experiments on two popular datasets, \ie, PASCAL-$5^i$~\cite{shaban2017one} and COCO-$20^i$~\cite{nguyen2019feature}. Since our method is model-agnostic, we choose two typical FSS methods, \ie, PFENet~\cite{tian2020prior} and BAM~\cite{lang2022learning}, as the baseline model to combine with our IPE.

\subsection{Datasets}
PASCAL-$5^i$~\cite{shaban2017one} and COCO-$20^i$~\cite{nguyen2019feature} are two popular datasets for FSS evaluations. Different methods have slight differences in the usage of the datasets, especially in the number of samples for testing, and our experiments follow the settings of PFENet. For comparing fairly with existing FSS methods, all extra data used by IPE are the unlabeled images belonging to the training set of ImageNet-1k~\cite{deng2009imagenet}. Note that almost all existing FSS methods use the supervisedly pre-trained model on ImageNet-1k as initialization for encoders while we explore the feasibility of employing an unsupervisedly pre-trained model as initialization. 

PASCAL-$5^i$~\cite{shaban2017one} is based on extended PASCAL VOC which is constructed by PASCAL VOC 2012~\cite{everingham2010pascal} and extra annotations from SBD~\cite{hariharan2011semantic}. PASCAL-$5^i$ evenly divides the 20 classes of extended PASCAL VOC into 4 folds (5 classes per fold) to execute cross-validation, namely, when one fold is selected as novel classes for the test, other remaining folds are used as base classes for training.

COCO-$20^i$~\cite{nguyen2019feature} is based on MS COCO~\cite{lin2014microsoft}, which is more challenging than PASCAL VOC. These challenges mainly come from more complex scenarios and more classes. COCO-$20^i$ evenly divides the 80 classes of MS COCO into 4 folds (20 classes per fold) to execute cross-validation, just like PASCAL-$5^i$.

\subsection{Evaluation metrics}
There are two metrics to evaluate FSS methods: Mean intersection over union (mIoU) and foreground-background intersection over union (FB-IoU). In our experiments, we choose both the mIoU and the FB-IoU as our evaluation metrics. 
The mIoU is the average of IoU over all the classes:
\begin{equation}
\mathrm{mIoU}=\frac{1}{N} \sum_{n=1}^{N} \mathrm{IoU}_{n},
\end{equation}
where N is the total number of novel classes and the IoU of each class is calculated as
\begin{equation}
\mathrm{IoU}=\frac{\mathrm{TP}}{\mathrm{TP}+\mathrm{FP}+\mathrm{FN}},
\end{equation}
where TP, FP and FN denote the number of true positive, false
positive and false negative pixels of the prediction, respectively.

The FB-IoU is the average of the foreground IoU and background IoU:
\begin{equation}
\mathrm{FB}\operatorname{-}\mathrm{IoU}=\frac{\mathrm{1}}{\mathrm{2}}(\mathrm{IoU_F} + \mathrm{IoU_B}),
\end{equation}
where the subscript $\mathrm{F}$ represents foreground and the subscript $\mathrm{B}$ represents background.

\subsection{Implementation details}
The encoders of PFENet~\cite{tian2020prior} and BAM~\cite{lang2022learning} are both ResNet-50~\cite{he2016deep}, which is usually supervisedly pre-trained on ImageNet-1K. We also explore the feasibility of utilizing an unsupervisedly pre-trained model provided by DenseCL~\cite{wang2021dense}. The batch size of extra episodes is four times the batch size of original inputs, and the image size of extra episodes is 225$\times$225.

In PGM, the feature extractor $\mathcal{H}$ is the unsupervisedly pre-trained ResNet-50~\cite{he2016deep}, whose parameters are from DenseCL~\cite{wang2021dense}. All unlabeled images are resized to the shape of 672$\times$672 before sending to the feature extractor. The cluster number $m$ and $n$ are $3$ and $5$, respectively. The pseudo-label number $t$ is $2$.

In EGM, we serially use \verb|RandomResizedCrop|, \verb|RandomRotation|, \verb|ColorJitter|, \verb|RandomGrayscale|, \verb|GaussianBlur| and \verb|RandomHorizontalFlip| with probabilities of 1.0, 0.4, 0.8, 0.2, 0.8 and 0.5 as various transformations.

\subsection{Results and analysis}
\label{section:result}

\subsubsection{PASCAL-$5^i$}

The 1-shot and 5-shot results on PASCAL-$5^i$~\cite{shaban2017one} are shown in Tables \ref{tab:pascal1} and \ref{tab:pascal5}, respectively. Under the 1-shot setting, after integrating with our IPE, PFENet achieves improvements up to 3.57\% and 2.17\% in terms of mIoU and FB-IoU, respectively. Moreover, when incorporating our IPE with a current strong baseline BAM, a promising result can be also obtained {with the improvements of 0.8\% in terms of both mIoU and FB-IoU.}

Besides, if the encoder of PFENet is initialized with an unsupervisedly pre-trained model, PFENet with IPE also outperforms PFENet with a larger improvement of up to 5.08\% in terms of FB-IoU. Interestingly, the unsupervisedly pre-trained PFENet with IPE even outperforms the supervisedly pre-trained PFENet without IPE, achieving improvements up to 1.77\% and 1.5\% in terms of mIoU and FB-IoU. We believe this is because our method unearths the semantic information hidden in the unlabeled data and brings this relatively complete supervision from abundant data to FSS models. All these results show that our IPE is simple but effective.

\begin{table}[h]
 \centering
 \setlength{\tabcolsep}{0.5mm}
 \caption{1-Shot mIoU and FB-IoU results on  PASCAL-$5^i$. All models in this table use ResNet-50 as their encoder. `S. IN' represents the encoder is supervisedly pre-trained on ImageNet-1k, `U. IN' represents the encoder is unsupervisedly pre-trained on ImageNet-1k.}
 \begin{tabular}{@{}c|c|c|cccc|cc@{}}
\toprule[1pt]
 Method & S. IN & U. IN  & Fold-0 & Fold-1 & Fold-2 & Fold-3 & Mean 
 & FB-IoU
 \\
 \midrule[1pt]
    PPNet\cite{liu2020part}&$\checkmark$& &48.58&60.58&55.71&46.47&52.84
    & -
    \\
    CANet~\cite{zhang2019canet}&$\checkmark$& & 52.50& 65.90& 51.30& 51.90& 55.40
    & 66.20
    \\
    RPMM~\cite{yang2020prototype}&$\checkmark$& & 55.15&66.91& 52.61 &50.68 &56.34
    & -
    \\
    ASGNet~\cite{li2021adaptive}&$\checkmark$& &58.84&67.86&56.79& 53.66&59.29 
    & 69.20
    \\
    RePRI~\cite{boudiaf2021few}&$\checkmark$& &59.80 &68.30 & 62.10 & 48.50 & 59.70
    & - 
    \\
    SCL~\cite{zhang2021self}&$\checkmark$& &63.00& 70.00& 56.50& 57.70& 61.80
    & 71.90
    \\
    HSNet~\cite{min2021hypercorrelation}& $\checkmark$& & 64.30 & 70.70 & 60.30 & 60.50 & 64.00 
     & 76.70  
     \\
     \midrule[1pt]
    PFENet~\cite{tian2020prior}& $\checkmark$ & &61.70 & 69.50 & 55.40 & 56.30 & 60.80
    & 73.30
    \\
    PFENet w/ \textbf{IPE}& $\checkmark$&  & 65.31 & 70.83 & 61.10 &  60.26  & 64.37 & 75.47
    \\
    BAM~\cite{lang2022learning}& $\checkmark$ & &68.97 & \textbf{73.59} & 67.55 & 61.13 & 67.81
    & 79.71
    \\
    BAM w/ \textbf{IPE}& $\checkmark$&  &\textbf{70.26} & 73.34 &\textbf{68.24} & \textbf{62.61} & \textbf{68.61} 
    &\textbf{80.51}
    \\ 
     \hline
    PFENet& &$\checkmark$ & 59.50 & \textbf{68.75} & 55.81 & 53.02 & 59.27 
    &  69.72
    \\
    PFENet w/ \textbf{IPE}& &$\checkmark$  &\textbf{62.26} & 67.78 &\textbf{64.22} & \textbf{56.02} & \textbf{62.57}
    & \textbf{74.80}
    \\

\bottomrule[1pt]
\end{tabular}
 \label{tab:pascal1}
\end{table}
\begin{table}[h]
 \centering
 \setlength{\tabcolsep}{0.5mm}
 \caption{5-Shot mIoU and FB-IoU results on  PASCAL-$5^i$. All models in this table use ResNet-50 as their encoder. All encoders are supervisedly pre-trained on ImageNet-1k.}
 \begin{tabular}{@{}c|cccc|cc@{}}
\toprule[1pt]
 Method & Fold-0 & Fold-1 & Fold-2 & Fold-3 & Mean 
 & FB-IoU
 \\
 \midrule[1pt]
    PPNet\cite{liu2020part}& 58.85 &68.28 &66.77& 57.98& 62.97 & -
    \\
    CANet~\cite{zhang2019canet} & 55.50& 67.80& 51.90& 53.20& 57.10& 69.60
    \\
    RPMM~\cite{yang2020prototype} & 56.28& 67.34& 54.52& 51.00& 57.30
    & -
    \\
    ASGNet~\cite{li2021adaptive} &63.66 & 70.55 & 64.17 & 57.38  &63.94 &74.2
    \\
    RePRI~\cite{boudiaf2021few} &64.60& 71.40& 71.10& 59.30& 66.60
    & - 
    \\
    SCL~\cite{zhang2021self} &64.50& 70.90& 57.30& 58.70& 62.90& 72.80
    \\
    HSNet~\cite{min2021hypercorrelation}& 70.30& 73.20& 67.40& 67.10& 69.50& 80.60
     \\
     \midrule[1pt]
    PFENet~\cite{tian2020prior} &63.10& 70.70& 55.80& 57.90& 61.90& 73.90
    \\
    PFENet w/ \textbf{IPE} & 68.11 & 73.37 & 73.88 &  68.64  & 71.00 & 81.63
    \\
    BAM~\cite{lang2022learning}&70.59& 75.05& 70.79& 67.20& 70.91
    & 82.18
    \\
    BAM w/ \textbf{IPE}&\textbf{72.60} & \textbf{75.22} & \textbf{73.50} & \textbf{68.42} & \textbf{72.44}
    & \textbf{83.46}
    \\ 

\bottomrule[1pt]
\end{tabular}
 \label{tab:pascal5}
\end{table}

\newpage
Under the 5-shot setting of PASCAL-$5^i$~\cite{shaban2017one}, similar conclusions can be obtained that both PFENet and BAM achieve better results when inserting our IPE. The PFENet with IPE reaches significant improvements up to 9.1\% and 7.73\% in terms of mIoU and FB-IoU, respectively. {Moreover, when combining our IPE with a strong baseline BAM, it also achieves further improvements up to 1.53\% and 1.28\% in terms of mIoU and FB-IoU.}

\begin{table}[htbp]
 \centering
  \setlength{\tabcolsep}{0.5mm}
 \caption{1-Shot mIoU and FB-IoU results on COCO-$20^i$. All models in this table use ResNet-50 as their encoder. `S. IN' represents the encoder is supervisedly pre-trained on ImageNet-1k, `U. IN' represents the encoder is unsupervisedly pre-trained on ImageNet-1k.}
 \begin{tabular}{@{}c|c|c|cccc|cc@{}}
\toprule[1pt]
 Method & S. IN & U. IN  & Fold-0 & Fold-1 & Fold-2 & Fold-3 & Mean 
 & FB-IoU
 \\
 \midrule[1pt]
    PPNet\cite{liu2020part}&$\checkmark$& &28.09&30.84& 29.49& 27.70 &29.03 
    &- 
    \\
   
    RPMM~\cite{yang2020prototype}&$\checkmark$&&29.53& 36.82 &28.94& 27.02& 30.58 
    &-  
    \\
    ASGNet~\cite{li2021adaptive}&$\checkmark$& & - & - &-  &-  & 34.56 
    &60.39 
    \\
    RePRI~\cite{boudiaf2021few}&$\checkmark$&&31.20& 38.10& 33.30& 33.00& 34.00 
    &-  
    \\
    HSNet~\cite{min2021hypercorrelation}&$\checkmark$&  &36.30& 43.10& 38.70& 38.70& 39.20 
    & 68.20
    \\
     \midrule[1pt]
    PFENet~\cite{tian2020prior}&$\checkmark$& &37.42& 41.08& 37.93& 35.96& 38.10 
    & 61.45
    \\
    PFENet w/ \textbf{IPE}&$\checkmark$& & 39.91 & 44.82 & 41.32 &  39.69 & 41.43
    & 64.88
    \\
    {BAM~\cite{lang2022learning}}& {$\checkmark$} & &{41.89}& {51.30}& {48.96}& {45.91}& {47.02 }   & {71.74}
    \\
    {BAM w/ \textbf{IPE}}& {$\checkmark$}&  &  {\textbf{42.78}} & {\textbf{51.90}} & {\textbf{49.43}} &  {\textbf{46.08}} & {\textbf{47.55}}
    & {\textbf{72.53}}
    \\

\hline
    PFENet& &$\checkmark$& 33.88 & 38.34 & $35.20$ & 34.85 & 35.57 
    &  61.00
    \\
    PFENet w/ \textbf{IPE}& &$\checkmark$& \textbf{36.74} & \textbf{40.62} & \textbf{37.20} &  \textbf{37.07} & \textbf{37.91}
    & \textbf{62.74}
    \\
\bottomrule[1pt]
\end{tabular}
 \label{tab:coco1}
\end{table}

\begin{table}[htbp]
 \centering
  \setlength{\tabcolsep}{0.5mm}
 \caption{5-Shot mIoU and FB-IoU results on COCO-$20^i$. All models in this table use ResNet-50 as their encoder. All encoders are supervisedly pre-trained on ImageNet-1k.}
 \begin{tabular}{@{}c|cccc|cc@{}}
\toprule[1pt]
 Method & Fold-0 & Fold-1 & Fold-2 & Fold-3 & Mean 
 & FB-IoU
 \\
 \midrule[1pt]

    PPNet\cite{liu2020part} &38.97& 40.81& 37.07& 37.28 &38.53
    &- 
    \\
   
    RPMM~\cite{yang2020prototype}& 33.82& 41.96 &32.99& 33.33& 35.52 
    &-  
    \\
    ASGNet~\cite{li2021adaptive} & - & - &-  &-  & 42.48 
    &66.96
    \\
    RePRI~\cite{boudiaf2021few}&39.30& 45.40& 39.70& 41.80& 41.60
    &-  
    \\
    HSNet~\cite{min2021hypercorrelation}&43.30& 51.30& 48.20& 45.00& 46.90& 70.70
    \\

     \midrule[1pt]
    PFENet~\cite{tian2020prior}&36.50& 43.30& 37.80& 38.40& 39.00
    & -
    \\
    PFENet w/ \textbf{IPE}& 44.86 & 51.47 & \textbf{50.67} & 45.92 & 48.23
    & 71.00
    
    \\
    {BAM~\cite{lang2022learning}}&{47.61}&{53.11}& {49.30}& {47.74}& {49.44}
    & {73.24}
    \\
    {BAM w/ \textbf{IPE}}& {\textbf{48.77}} & {\textbf{56.05}} & {50.59} & {\textbf{48.13}} & {\textbf{50.89}}
    & {\textbf{73.66}}
    \\

\bottomrule[1pt]
\end{tabular}
 \label{tab:coco5}
\end{table}

\subsubsection{COCO-$20^i$}

The 1-shot and 5-shot results on COCO-$20^i$~\cite{shaban2017one} are shown in Tables \ref{tab:coco1} and \ref{tab:coco5}, respectively. Under the 1-shot setting, after combining with our IPE, the mIoU of PFENet exhibits improvements up to 3.33\% and 2.34\% with the supervisedly and unsupervisedly pre-trained encoder, respectively. The unsupervisedly pre-trained PFENet with our IPE even achieves a comparable result to the supervisedly pre-trained PFENet without IPE. {Besides, we conduct the experiments on BAM with our IPE. It can be seen that our method also outperforms BAM. More unlabelled data may further improve the accuracy, which is an interesting direction for further study. }

Under the 5-shot setting of COCO-$20^i$, similar conclusion can be obtained that the PFENet with IPE reaches a significant improvement up to 9.23\% of mIoU{, and the BAM with IPE reaches the improvements up to 1.45\% of mIoU. All these results demonstrates the effectiveness of our IPE leveraging unlabelled data for boosting FSS.} 

\subsubsection{Result Visualization}

\begin{figure*}[htbp]
 \centering
 \includegraphics[width=1\linewidth]{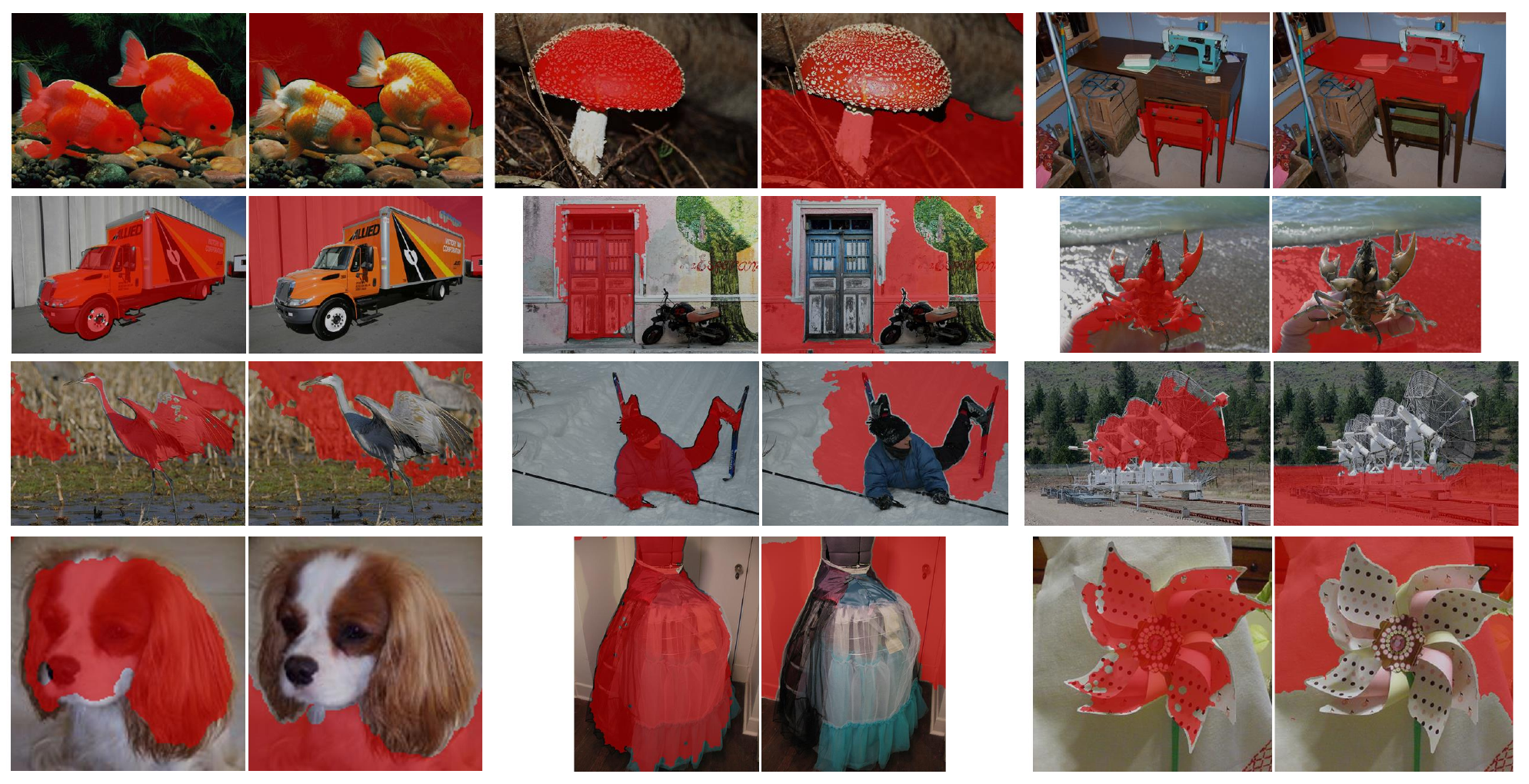}
 \caption{Some pseudo-labels generated by PGM from ImageNet-1k.}
 \label{fig:visualize}
\end{figure*}

In Figure \ref{fig:visualize}, we show a batch of pseudo-labels generated by PGM. Although the accuracy of some pseudo-labels is not very high, these pseudo-labels are good enough to provide extra information for semantics. Besides, the boundary of these labels is also clear enough with the help of dense-CRF. 
In general, the pseudo-labels generated by PGM significantly improve the generalization ability of the FSS models.

\begin{figure*}[htbp]
 \centering
 \includegraphics[width=1\linewidth]{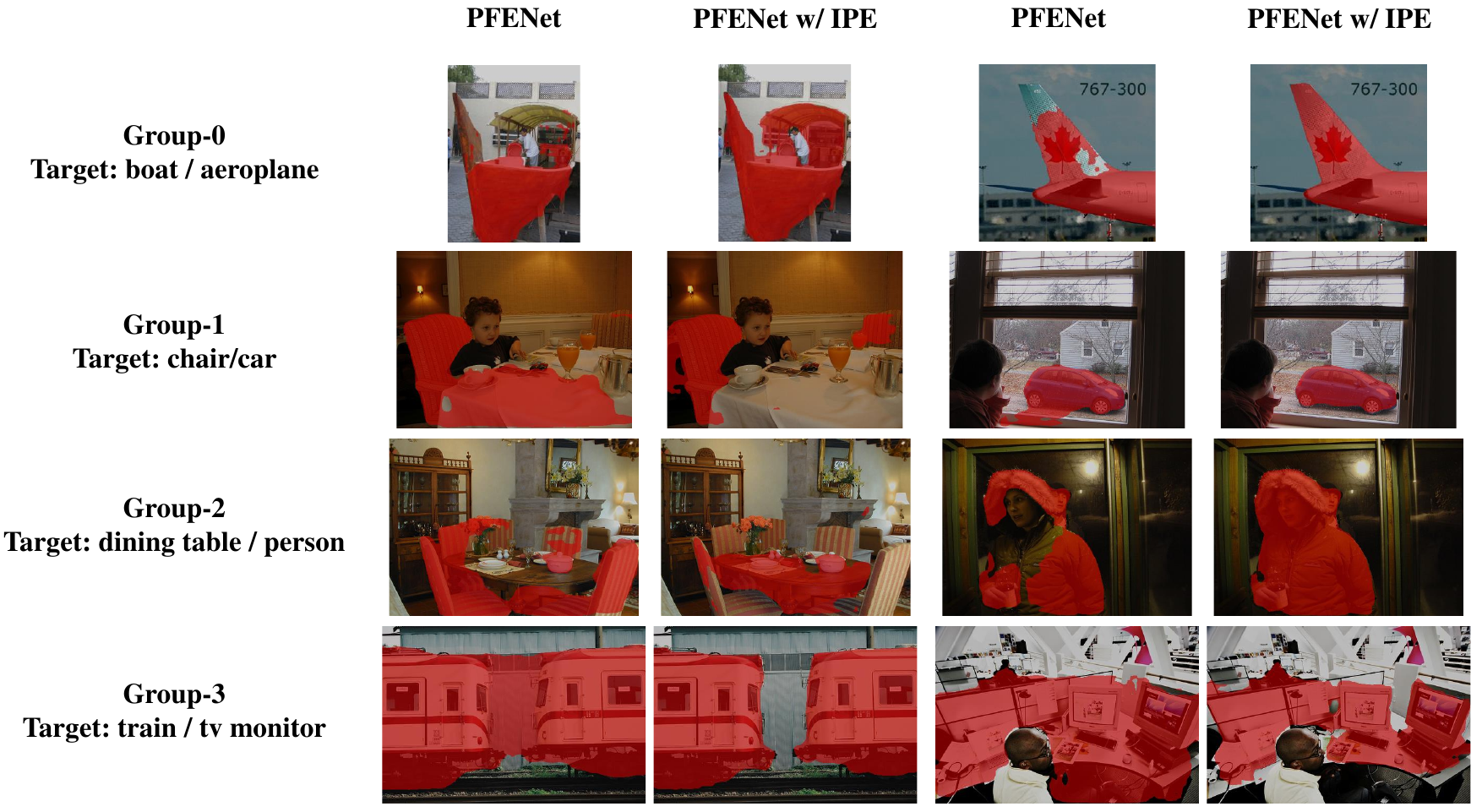}
 \caption{Some predictions generated by PFENet on PASCAL-$5^i$ under the 1-shot setting with and without IPE. Each pair of pictures shows the effectiveness of IPE in every group.}
 \label{fig:visualize_2}
\end{figure*}

In Figure \ref{fig:visualize_2}, we show a batch of predictions generated by PFENet on PASCAL-$5^i$ under the 1-shot setting with or without IPE. Specifically, in group-0, our IPE helps to achieve more complete segmentation results of ``boat'' and ``aeroplane''. In group-1, the PFENet with IPE well segments the ``chair'' out correctly; and in group-2, our IPE corrects the original PFENet segmentation error of treating the base class ``chair'' as the novel class ``dining table''. Besides, for group-3, our IPE can well remove the false positive regions on the background for ``train'' and ``tv monitor''. All these results indicate that our IPE can improve the segmentation accuracy with more complete coverage of foreground objects and fewer false positive regions on the background.

\section{Limitation}
Although our approach allows incorporating additional unlabeled data into the training of FSS models to improve their generalization ability and accuracy, it still has its limitations. The main limitation of our IPE may lie in struggling on dealing with unlabeled extra data under complex scenes. Since the image of complex scene has many objects of different classes, it is hard for PGM to generate accurate pseudo labels with spectral clustering. Fortunately, we can easily collect images of limited objects, like ImageNet, which can significantly boost the accuracy of few-shot segmentation.In the future, we will explore new metrics to measure the complexity of the image to flexibly select appropriate samples or resort to multi-model self-supervision learning for better understanding images under complex scenes.

\section{Conclusion}
We propose a novel model-agnostic method, named Image to Pseudo-Episode (IPE), to boost the performance of FSS methods by leveraging extra unlabeled data. Through generating pseudo-labels from unlabeled images by PGM and generating pseudo-episodes from the pseudo-labeled images by EGM, our IPE provides a bridge to introduce the extra information from unlabeled data into FSS models, thereby improving the generalization ability of these models.. With the help of IPE, both PFENet~\cite{tian2020prior} and BAM\cite{lang2022learning} achieve better results on PASCAL-$5^i$ and COCO-$20^i$ in terms of qualitative and quantitative evaluations.

\newpage

\bibliography{mybibfile}

\begin{thebibliography}{10}
\expandafter\ifx\csname url\endcsname\relax
  \def\url#1{\texttt{#1}}\fi
\expandafter\ifx\csname urlprefix\endcsname\relax\def\urlprefix{URL }\fi
\expandafter\ifx\csname href\endcsname\relax
  \def\href#1#2{#2} \def\path#1{#1}\fi

\bibitem{he2016deep}
K.~He, X.~Zhang, S.~Ren, J.~Sun, Deep residual learning for image recognition,
  in: Proceedings of the IEEE conference on computer vision and pattern
  recognition, 2016, pp. 770--778.

\bibitem{he2017mask}
K.~He, G.~Gkioxari, P.~Doll{\'a}r, R.~Girshick, Mask r-cnn, in: Proceedings of
  the IEEE international conference on computer vision, 2017, pp. 2961--2969.

\bibitem{chen2018encoder}
L.-C. Chen, Y.~Zhu, G.~Papandreou, F.~Schroff, H.~Adam, Encoder-decoder with
  atrous separable convolution for semantic image segmentation, in: Proceedings
  of the European conference on computer vision (ECCV), 2018, pp. 801--818.

\bibitem{chen2020simple}
T.~Chen, S.~Kornblith, M.~Norouzi, G.~Hinton, A simple framework for
  contrastive learning of visual representations, in: International conference
  on machine learning, PMLR, 2020, pp. 1597--1607.

\bibitem{wang2021dense}
X.~Wang, R.~Zhang, C.~Shen, T.~Kong, L.~Li, Dense contrastive learning for
  self-supervised visual pre-training, in: Proceedings of the IEEE/CVF
  Conference on Computer Vision and Pattern Recognition, 2021, pp. 3024--3033.

\bibitem{kolesnikov2016seed}
A.~Kolesnikov, C.~H. Lampert, Seed, expand and constrain: Three principles for
  weakly-supervised image segmentation, in: European conference on computer
  vision, Springer, 2016, pp. 695--711.

\bibitem{saleh2016built}
F.~Saleh, M.~S. Aliakbarian, M.~Salzmann, L.~Petersson, S.~Gould, J.~M.
  Alvarez, Built-in foreground/background prior for weakly-supervised semantic
  segmentation, in: European conference on computer vision, Springer, 2016, pp.
  413--432.

\bibitem{zhang2019canet}
C.~Zhang, G.~Lin, F.~Liu, R.~Yao, C.~Shen, Canet: Class-agnostic segmentation
  networks with iterative refinement and attentive few-shot learning, in:
  Proceedings of the IEEE/CVF Conference on Computer Vision and Pattern
  Recognition, 2019, pp. 5217--5226.

\bibitem{min2021hypercorrelation}
J.~Min, D.~Kang, M.~Cho, Hypercorrelation squeeze for few-shot segmentation,
  in: Proceedings of the IEEE/CVF International Conference on Computer Vision,
  2021, pp. 6941--6952.

\bibitem{tian2020prior}
Z.~Tian, H.~Zhao, M.~Shu, Z.~Yang, R.~Li, J.~Jia, Prior guided feature
  enrichment network for few-shot segmentation, IEEE transactions on pattern
  analysis and machine intelligence.

\bibitem{yang2020prototype}
B.~Yang, C.~Liu, B.~Li, J.~Jiao, Q.~Ye, Prototype mixture models for few-shot
  semantic segmentation, in: European Conference on Computer Vision, Springer,
  2020, pp. 763--778.

\bibitem{li2021adaptive}
G.~Li, V.~Jampani, L.~Sevilla-Lara, D.~Sun, J.~Kim, J.~Kim, Adaptive prototype
  learning and allocation for few-shot segmentation, in: Proceedings of the
  IEEE/CVF Conference on Computer Vision and Pattern Recognition, 2021, pp.
  8334--8343.

\bibitem{shi2022dense}
X.~Shi, D.~Wei, Y.~Zhang, D.~Lu, M.~Ning, J.~Chen, K.~Ma, Y.~Zheng, Dense
  cross-query-and-support attention weighted mask aggregation for few-shot
  segmentation, in: European Conference on Computer Vision, Springer, 2022, pp.
  151--168.

\bibitem{johnander2022dense}
J.~Johnander, J.~Edstedt, M.~Felsberg, F.~S. Khan, M.~Danelljan, Dense gaussian
  processes for few-shot segmentation, in: European Conference on Computer
  Vision, Springer, 2022, pp. 217--234.

\bibitem{zhang2022feature}
J.-W. Zhang, Y.~Sun, Y.~Yang, W.~Chen, Feature-proxy transformer for few-shot
  segmentation, Advances in Neural Information Processing Systems 35 (2022)
  6575--6588.

\bibitem{simonyan2014very}
K.~Simonyan, A.~Zisserman, Very deep convolutional networks for large-scale
  image recognition, arXiv preprint arXiv:1409.1556.

\bibitem{deng2009imagenet}
J.~Deng, W.~Dong, R.~Socher, L.-J. Li, K.~Li, L.~Fei-Fei, Imagenet: A
  large-scale hierarchical image database, in: 2009 IEEE conference on computer
  vision and pattern recognition, Ieee, 2009, pp. 248--255.

\bibitem{shaban2017one}
A.~Shaban, S.~Bansal, Z.~Liu, I.~Essa, B.~Boots, One-shot learning for semantic
  segmentation, arXiv preprint arXiv:1709.03410.

\bibitem{fiedler1973algebraic}
M.~Fiedler, Algebraic connectivity of graphs, Czechoslovak mathematical journal
  23~(2) (1973) 298--305.

\bibitem{krahenbuhl2011efficient}
P.~Kr{\"a}henb{\"u}hl, V.~Koltun, Efficient inference in fully connected crfs
  with gaussian edge potentials, Advances in neural information processing
  systems 24.

\bibitem{he2020momentum}
K.~He, H.~Fan, Y.~Wu, S.~Xie, R.~Girshick, Momentum contrast for unsupervised
  visual representation learning, in: Proceedings of the IEEE/CVF conference on
  computer vision and pattern recognition, 2020, pp. 9729--9738.

\bibitem{grill2020bootstrap}
J.-B. Grill, F.~Strub, F.~Altch{\'e}, C.~Tallec, P.~Richemond, E.~Buchatskaya,
  C.~Doersch, B.~Avila~Pires, Z.~Guo, M.~Gheshlaghi~Azar, et~al., Bootstrap
  your own latent-a new approach to self-supervised learning, Advances in
  Neural Information Processing Systems 33 (2020) 21271--21284.

\bibitem{lin2014microsoft}
T.-Y. Lin, M.~Maire, S.~Belongie, J.~Hays, P.~Perona, D.~Ramanan,
  P.~Doll{\'a}r, C.~L. Zitnick, Microsoft coco: Common objects in context, in:
  European conference on computer vision, Springer, 2014, pp. 740--755.

\bibitem{nguyen2019feature}
K.~Nguyen, S.~Todorovic, Feature weighting and boosting for few-shot
  segmentation, in: Proceedings of the IEEE/CVF International Conference on
  Computer Vision, 2019, pp. 622--631.

\bibitem{lang2022learning}
C.~Lang, G.~Cheng, B.~Tu, J.~Han, Learning what not to segment: A new
  perspective on few-shot segmentation, in: Proceedings of the IEEE/CVF
  Conference on Computer Vision and Pattern Recognition, 2022, pp. 8057--8067.

\bibitem{lin2017refinenet}
G.~Lin, A.~Milan, C.~Shen, I.~Reid, Refinenet: Multi-path refinement networks
  for high-resolution semantic segmentation, in: Proceedings of the IEEE
  conference on computer vision and pattern recognition, 2017, pp. 1925--1934.

\bibitem{zhao2017pyramid}
H.~Zhao, J.~Shi, X.~Qi, X.~Wang, J.~Jia, Pyramid scene parsing network, in:
  Proceedings of the IEEE conference on computer vision and pattern
  recognition, 2017, pp. 2881--2890.

\bibitem{ronneberger2015u}
O.~Ronneberger, P.~Fischer, T.~Brox, U-net: Convolutional networks for
  biomedical image segmentation, in: International Conference on Medical image
  computing and computer-assisted intervention, Springer, 2015, pp. 234--241.

\bibitem{chen2017rethinking}
L.-C. Chen, G.~Papandreou, F.~Schroff, H.~Adam, Rethinking atrous convolution
  for semantic image segmentation, arXiv preprint arXiv:1706.05587.

\bibitem{orvsic2021efficient}
M.~Or{\v{s}}i{\'c}, S.~{\v{S}}egvi{\'c}, Efficient semantic segmentation with
  pyramidal fusion, Pattern Recognition 110 (2021) 107611.

\bibitem{long2015fully}
J.~Long, E.~Shelhamer, T.~Darrell, Fully convolutional networks for semantic
  segmentation, in: Proceedings of the IEEE conference on computer vision and
  pattern recognition, 2015, pp. 3431--3440.

\bibitem{chen2017deeplab}
L.-C. Chen, G.~Papandreou, I.~Kokkinos, K.~Murphy, A.~L. Yuille, Deeplab:
  Semantic image segmentation with deep convolutional nets, atrous convolution,
  and fully connected crfs, IEEE transactions on pattern analysis and machine
  intelligence 40~(4) (2017) 834--848.

\bibitem{chen2019closer}
W.-Y. Chen, Y.-C. Liu, Z.~Kira, Y.-C.~F. Wang, J.-B. Huang, A closer look at
  few-shot classification, arXiv preprint arXiv:1904.04232.

\bibitem{snell2017prototypical}
J.~Snell, K.~Swersky, R.~Zemel, Prototypical networks for few-shot learning,
  Advances in neural information processing systems 30.

\bibitem{garcia2017few}
V.~Garcia, J.~Bruna, Few-shot learning with graph neural networks, arXiv
  preprint arXiv:1711.04043.

\bibitem{koch2015siamese}
G.~Koch, R.~Zemel, R.~Salakhutdinov, et~al., Siamese neural networks for
  one-shot image recognition, in: ICML deep learning workshop, Vol.~2, Lille,
  2015, p.~0.

\bibitem{oreshkin2018tadam}
B.~Oreshkin, P.~Rodr{\'\i}guez~L{\'o}pez, A.~Lacoste, Tadam: Task dependent
  adaptive metric for improved few-shot learning, Advances in neural
  information processing systems 31.

\bibitem{vinyals2016matching}
O.~Vinyals, C.~Blundell, T.~Lillicrap, D.~Wierstra, et~al., Matching networks
  for one shot learning, Advances in neural information processing systems 29.

\bibitem{huang2021local}
H.~Huang, Z.~Wu, W.~Li, J.~Huo, Y.~Gao, Local descriptor-based multi-prototype
  network for few-shot learning, Pattern Recognition 116 (2021) 107935.

\bibitem{dong2018few}
N.~Dong, E.~P. Xing, Few-shot semantic segmentation with prototype learning.,
  in: BMVC, Vol.~3, 2018.

\bibitem{lin2013network}
M.~Lin, Q.~Chen, S.~Yan, Network in network, arXiv preprint arXiv:1312.4400.

\bibitem{zhang2020sg}
X.~Zhang, Y.~Wei, Y.~Yang, T.~S. Huang, Sg-one: Similarity guidance network for
  one-shot semantic segmentation, IEEE Transactions on Cybernetics 50~(9)
  (2020) 3855--3865.

\bibitem{sun2022singular}
Y.~Sun, Q.~Chen, X.~He, J.~Wang, H.~Feng, J.~Han, E.~Ding, J.~Cheng, Z.~Li,
  J.~Wang, Singular value fine-tuning: Few-shot segmentation requires
  few-parameters fine-tuning, Advances in Neural Information Processing Systems
  35 (2022) 37484--37496.

\bibitem{zhang2021few}
G.~Zhang, G.~Kang, Y.~Yang, Y.~Wei, Few-shot segmentation via cycle-consistent
  transformer, Advances in Neural Information Processing Systems 34 (2021)
  21984--21996.

\bibitem{zhang2021self}
B.~Zhang, J.~Xiao, T.~Qin, Self-guided and cross-guided learning for few-shot
  segmentation, in: Proceedings of the IEEE/CVF Conference on Computer Vision
  and Pattern Recognition, 2021, pp. 8312--8321.

\bibitem{xie2021detco}
E.~Xie, J.~Ding, W.~Wang, X.~Zhan, H.~Xu, P.~Sun, Z.~Li, P.~Luo, Detco:
  Unsupervised contrastive learning for object detection, in: Proceedings of
  the IEEE/CVF International Conference on Computer Vision, 2021, pp.
  8392--8401.

\bibitem{calinski1974dendrite}
T.~Cali{\'n}ski, J.~Harabasz, A dendrite method for cluster analysis,
  Communications in Statistics-theory and Methods 3~(1) (1974) 1--27.

\bibitem{everingham2010pascal}
M.~Everingham, L.~Van~Gool, C.~K. Williams, J.~Winn, A.~Zisserman, The pascal
  visual object classes (voc) challenge, International journal of computer
  vision 88~(2) (2010) 303--338.

\bibitem{hariharan2011semantic}
B.~Hariharan, P.~Arbel{\'a}ez, L.~Bourdev, S.~Maji, J.~Malik, Semantic contours
  from inverse detectors, in: 2011 international conference on computer vision,
  IEEE, 2011, pp. 991--998.

\bibitem{liu2020part}
Y.~Liu, X.~Zhang, S.~Zhang, X.~He, Part-aware prototype network for few-shot
  semantic segmentation, in: European Conference on Computer Vision, Springer,
  2020, pp. 142--158.

\bibitem{boudiaf2021few}
M.~Boudiaf, H.~Kervadec, Z.~I. Masud, P.~Piantanida, I.~Ben~Ayed, J.~Dolz,
  Few-shot segmentation without meta-learning: A good transductive inference is
  all you need?, in: Proceedings of the IEEE/CVF Conference on Computer Vision
  and Pattern Recognition, 2021, pp. 13979--13988.

\end{thebibliography}

\end{document}